\documentclass[9pt,twocolumn]{article}

\usepackage{microtype}
\usepackage{graphicx}
\usepackage{subfigure}
\usepackage{booktabs}       
\usepackage{paralist}       

\usepackage{hyperref}
\hypersetup{hyperfootnotes=false}
\usepackage[authoryear]{natbib}
\usepackage{amsmath}
\usepackage{amssymb}
\usepackage{mathtools}
\usepackage{amsthm}

\usepackage[T1]{fontenc}
\usepackage{lmodern}

\usepackage[capitalize,noabbrev]{cleveref}

\usepackage[textsize=tiny]{todonotes}

\theoremstyle{plain}
\newtheorem{theorem}{Theorem}[section]

\theoremstyle{definition}
\newtheorem{definition}[theorem]{Definition}

\theoremstyle{remark}

\title{From Directions to Cones: Exploring Multidimensional Representations of Propositional Facts in LLMs}

\author{%
  Stanley Yu\textsuperscript{*}, Vaidehi Bulusu\textsuperscript{*}, Oscar Yasunaga, Clayton Lau, \\
  Cole Blondin, Sean O’Brien, Kevin Zhu, Vasu Sharma\\
}

\date{}  

\makeatletter
\newcommand\blfootnote[1]{%
  \begingroup
    \renewcommand\thefootnote{}
    \hypersetup{draft}
    \footnote{#1}%
    \addtocounter{footnote}{-1}
  \endgroup
}
\makeatother

\begin{document}

\maketitle

\blfootnote{* Equal contribution. Correspondence to \texttt{stany@seas.upenn.edu}.}

\begin{abstract}
Large Language Models (LLMs) exhibit strong conversational abilities but often generate falsehoods. Prior work suggests that the truthfulness of simple propositions can be represented as a single linear direction in a model’s internal activations, but this may not fully capture its underlying geometry. In this work, we extend the concept cone framework, recently introduced for modeling refusal, to the domain of truth. We identify multi-dimensional cones that causally mediate truth-related behavior across multiple LLM families. Our results are supported by three lines of evidence: (i) causal interventions reliably flip model responses to factual statements, (ii) learned cones generalize across model architectures, and (iii) cone-based interventions preserve unrelated model behavior. These findings reveal the richer, multidirectional structure governing simple true/false propositions in LLMs and highlight concept cones as a promising tool for probing abstract behaviors.
\end{abstract}

\begin{figure}[!t]
\centering
\includegraphics[width=\columnwidth]{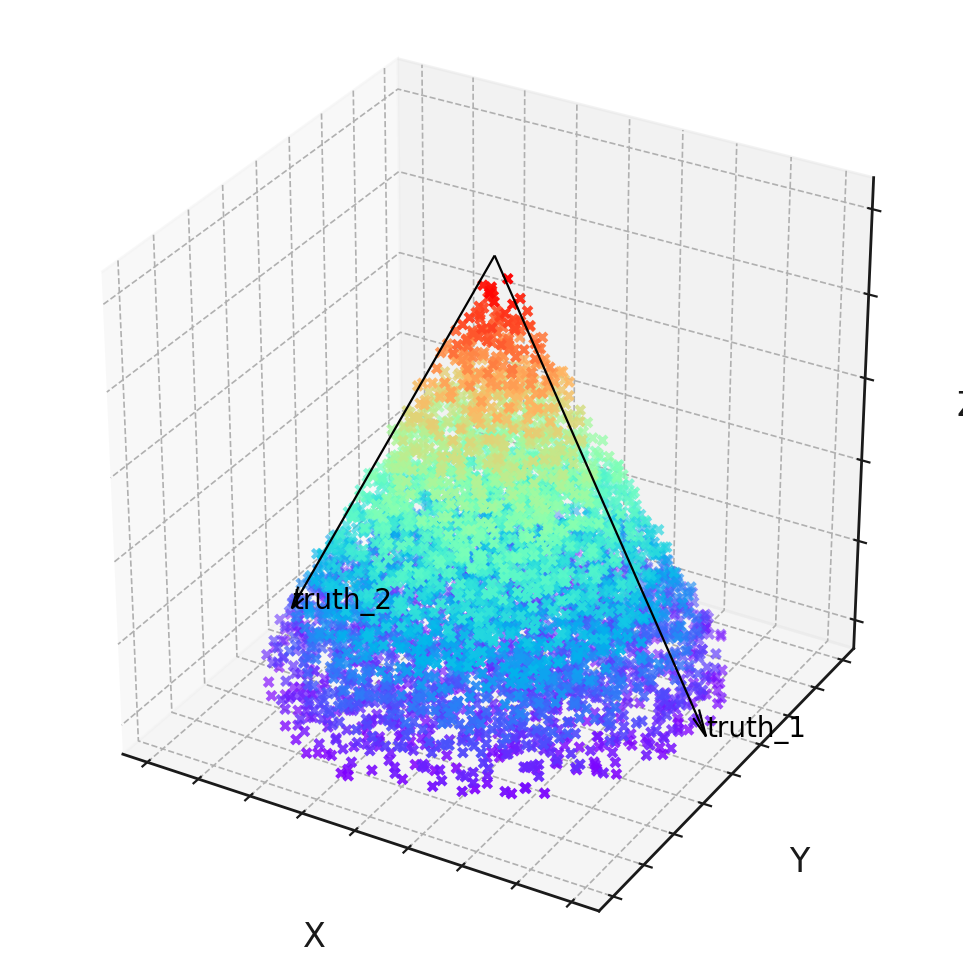}
\caption{Theoretical visualization of a 2D concept cone. All directions in the cone should causally mediate truthful behavior. Given a true propositional input (e.g., “Paris is the capital of France”), ablating along any basis vector of this cone disrupts the model’s ability to generate a truthful response.}
\label{fig:myfigure}
\end{figure}

\section{Introduction}
\label{introduction}
In recent years, Large Language Models (LLMs) have demonstrated remarkable capabilities across a wide range of natural language processing tasks, including machine translation, question answering, summarization, code generation, and dialogue systems \cite{brown2020language,raffel2020exploring,zhang2020pegasus,openai2023gpt4}. Despite their successes, these models remain largely “black boxes” with billions of parameters interacting in complex ways that evade straightforward analysis \cite{casper2024blackbox}. This presents challenges for ensuring alignment with human values and addressing vulnerabilities to adversarial attacks \cite{hendrycks2023overview, ngo2022alignment, hendrycks2022xrisk}. As these models are widely deployed in real-world applications, concerns about reliability and safety have driven a growing interest in model transparency \cite{openai2022chatgpt, olah2020zoom, nanda2023progress}. Specifically, identifying how and why specific linguistic or behavioral features are encoded within these models is one of the central questions for mechanistic interpretability research \cite{bereska2024mechanistic}.

To analyze the internal representations of LLMs, causal methods such as activation steering and directional ablation \cite{turner2024activation} are used to verify whether modifying specific internal directions leads to corresponding changes in model behavior \cite{panickssery2023steering, chen2024yesmen}. Together, probing and causal interventions have provided insight into how abstract features manifest in model representations.

Previous interpretability studies \cite{park2024linear, pmlr-v235-park24c} have revealed that many high-level features in LLMs correspond to linear directions in the representation space, such as time \cite{gurnee2023language}, truth \cite{marks2023geometry, azaria2023internal}, space \cite{gurnee2023language}, political perspective \cite{kim2025political}, and instruction-following \cite{heo2025instructions}. Other features such as sentiment \cite{tigges2023linear} and refusal \cite{arditi2024refusal} have also been shown to exist linearly, although through a different interpretability method known as difference-in-means (DIM). However, the underlying representations may be non-linear, and linear methods may only provide an approximation to more complex structures \cite{burger2024truth, hildebrandt2025refusal, engels2024not}. Recent work has developed more sophisticated non-linear frameworks and found multiple latent dimensions that capture fundamental high-level concepts, notably for refusal \cite{hildebrandt2025refusal, wollschlager2025geometry}.

With sparse autoencoders and concept cones, researchers have characterized multi-dimensional representations of abstract features \cite{cunningham2023sparse, liu2023cones, sharkey2023superposition}. Concept cones use a gradient-based search algorithm that, given candidate vectors, learns a specific behavior. Each vector is validated to causally influence the target concept through steering or ablation. This method extends the interpretability toolkit beyond linear assumptions by enabling both analysis and controlled intervention \cite{liu2023cones, wollschlager2025geometry}.

In this paper, we extend the concept cone framework to the domain of propositional fact, a subcategory of truthfulness, exploring how this property is internally represented by LLMs. Specifically, by applying this framework, we identify a multi-dimensional subspace whose basis vectors each contribute to the model’s ability to distinguish propositional true and false statements.

\section{Background}
\label{background}
\subsection{Transformers}
Decoder-only transformers \citep{liu2018generating} map input tokens $\mathbf{t} = (t_1, t_2, \ldots, t_n) \in \mathcal{V}^n$ to output probability distributions $\mathbf{y} = (\mathbf{y}_1, \mathbf{y}_2, \ldots, \mathbf{y}_n) \in \mathbb{R}^{n \times |\mathcal{V}|}$. Let $\mathbf{x}_i^{(l)}(\mathbf{t}) \in \mathbb{R}^{d_{\text{model}}}$ denote the residual stream activation of the token at position $i$ at the start of layer~$l$. \footnote{We shorten $\mathbf{x}_i^{(l)}(\mathbf{t})$ to $\mathbf{x}_i^{(l)}$ when the input $\mathbf{t}$ is clear from context or unimportant.} Each token's residual stream is initialized to its embedding $\mathbf{x}_i^{(1)} = \mathtt{Embed}(t_i)$, and then undergoes a series of transformations across $L$ layers. Each layer's transformation includes contributions from attention and MLP components:
\begin{align}
\tilde{\mathbf{x}}_i^{(l)} &= \mathbf{x}_i^{(l)} + \mathtt{Attn}^{(l)}(\mathbf{x}_{1:i}^{(l)}) \\
\mathbf{x}_i^{(l+1)} &= \tilde{\mathbf{x}}_i^{(l)} + \mathtt{MLP}^{(l)}(\tilde{\mathbf{x}}_i^{(l)}).
\end{align}
The final logits $\mathtt{logits}_i = \mathtt{Unembed}(\mathbf{x}_i^{(L+1)}) \in \mathbb{R}^{|\mathcal{V}|}$ are then transformed into probabilities over output tokens $\mathbf{y}_i = \mathtt{softmax}(\mathtt{logits}_i) \in \mathbb{R}^{|\mathcal{V}|}$.\footnote{This high-level description omits details such as positional embeddings and layer normalization.}

\subsection{Internal Representations of Truth}
\label{subsec: Internal Representations of Truth}

Recent work suggests that LLMs can encode factuality internally, even if their outputs does not always reflect it \cite{azaria2023internal}. Methods like linear probing and DIM have been used to identify directions in the activation space, often in the residual stream, that correlate with whether a statement is true or false \cite{marks2023geometry, burger2024truth}. We draw inspiration from these works by using labeled data sets of true and false English statements to investigate how the truth is geometrically embedded in the hidden states of the model. Similar to \citet{marks2023geometry} and \citet{burger2024truth}, we define truth as a specific operationalization: simple, unambiguous propositional statements that can be labeled as true or false.

Following \citet{wollschlager2025geometry}, who define refusal properties for vectors, we define analogous truth properties for vectors.

\begin{definition}
Truth Property
\begin{itemize}
   \item \textit{Monotonic Scaling:} when using the direction for activation addition/ablation\\ $\smash{\hat{\mathbf{x}}^{(l)}_{i} = \hat{\mathbf{x}}^{(l)}_i + \alpha \cdot \mathbf{r}}$, the model's probability of being more truthful should scale monotonically with $\alpha$. So, the percentage by which the model flips to the opposite answer (e.g. from no to yes) should scale monotonically with $\alpha$.
    \item \textit{Surgical Ablation} Ablating the truth direction through projection
\begin{align}
    \tilde{\mathbf{x}}_i^{(l)} \leftarrow \mathbf{x}_i^{(l)} - \hat{\mathbf{r}} \hat{\mathbf{r}}^{\intercal} \mathbf{x}_i^{(l)}. \label{eq:projection}
\end{align}
should cause the model to shift the answer from an initially true output to a false output.   
\end{itemize}
\end{definition}

\subsection{Model Interventions}
\label{subsec:model_interventions}

\subsubsection{Activation Addition}
Given a linear direction vector that represents a concept $\mathbf{r}^{(l)} \in \mathbb{R}^{d_{\text{model}}}$ extracted from layer $l$, we can use linear interventions such as addition and subtraction, scaled by some coefficient \(\alpha \in \mathbb{R}\), to modulate the strength of the corresponding feature in the activation space. For example, adding a learned truth vector to the activations shifts the representation toward regions of the activation space associated with truthful outputs.
\begin{align}
\mathbf{x}^{(l)'} \leftarrow \mathbf{x}^{(l)} + \alpha \cdot \mathbf{r}^{(l)}. \label{eq:activation_addition}
\end{align}
Note that for activation addition, we intervene only at layer $l$, and across all token positions.

\subsubsection{Directional Ablation}
To investigate the role of a direction $\hat{\mathbf{r}} \in \mathbb{R}^{d_{\text{model}}}$ in the model's computation, we can erase it from the model's representations using \emph{directional ablation} \citep{arditi2024refusal}.

Directional ablation subtracts the component along $\hat{\mathbf{r}}$ for every residual stream activation $\mathbf{x} \in \mathbb{R}^{d_{\text{model}}}$:
\begin{align}
    \mathbf{x}' \leftarrow \mathbf{x} - \hat{\mathbf{r}} \hat{\mathbf{r}}^{\intercal} \mathbf{x}. \label{eq:projection}
\end{align}
We perform this operation at every activation $\mathbf{x}_{i}^{(l)}$ and $\tilde{\mathbf{x}}_{i}^{(l)}$, across all layers $l$ and all token positions $i$.
This effectively prevents the model from ever representing this direction in its residual stream.

\subsection{Gradient-Based Methods}
\label{subsec: Gradient-Based Methods}

Gradient-based methods are a class of interpretability techniques that use gradients of model outputs with respect to internal activations to identify influential features or directions by revealing how small changes influence predictions. More recently, \citet{wollschlager2025geometry} have used gradients to steer model behavior: specific objectives, such as refusing unsafe inputs, can be encoded directly as loss functions. By optimizing a single vector that is added to or ablated from activations at specific layers, models can be guided toward target behaviors (e.g., safe refusals) while minimizing side effects on unrelated outputs. When applied to truthfulness, this framework enables precise, interpretable interventions and allows models to express truth-aligned responses without requiring full fine-tuning.

\subsection{Concept Cones}
\label{subsec: Concept Cones}

As described in \citet{wollschlager2025geometry}, given a set of orthonormal vectors $V = [v_1, v_2, \ldots,v_k] \in \mathbb{R}^{d_{model}\times k}$ a matrix whose columns are vectors each exhibit truth properties. The cone is the set of all nonnegative linear combinations of $$\mathcal{R}_N = \{\sum^k_{i=1} \lambda_iv_i \mid \lambda_i \geq 0\}\setminus\{0\}$$ All directions used in the cone correspond to the same truth concept. The constraint $\{\lambda_i \geq 0\}$ ensures that all directions within the cone consistently strengthen truth behavior.

\section{Methodology}
\label{Sec: Methodology}
To investigate the existence and structure of directions representing the notion of truthfulness in language models, we start with a linear-probe paradigm introduced by \citet{marks2023geometry},
we locate a linear direction in the residual stream by feeding the model raw factual statements and regressing on their ground-truth labels. We retain their definition of using factual statements that are simple, unambiguous and have topical diversity. 

We modify the following: instead of attaching a label offline, we ask the model to answer each statement with a binary ``Yes'' or ``No’’ and use that forced choice as the supervision signal.  This lets us treat the model’s own response distribution as a self-labeled probe target, enabling activation addition and ablation tests on the same forward pass and providing targets for our gradient descent approach.

\subsection{Setup for Truth Representation Discovery}
Each experiment involves prompting the LLM with a short factual statement and requesting a binary ``Yes'' or ``No'' response. We format the prompt using a system instruction to make it clear that the model should answer truthfully and concisely (See Appendix~\ref{system-prompts}  for all system prompts used). For example:

\begin{quote}
\texttt{\textcolor{blue}{System: Respond to the following statements with either ``Yes'' or ``No'' based on their factual accuracy.}} \\
\texttt{User: The Eiffel Tower is in Paris.} \\
\texttt{\textcolor{red}{Model: Yes ...}}
\end{quote}

We assume that for sufficiently capable base models, correct classification is achieved under normal conditions. Our goal is to test whether internal directions in activation space causally mediate this truthful behavior.

\subsection{Causal Interventions: Addition and Ablation}
Let $\mathbf{r}^{(l)} \in \mathbb{R}^{d_{\text{model}}}$ be a candidate direction vector associated with the concept of truth at layer $l$. We apply addition and ablation as follows:

\begin{itemize}
    \item \textbf{Directional Addition}: Given a false statement (where the base model typically outputs ``No''), we apply $\mathbf{r}^{(l)}$ additively to shift the model's behavior toward ``Yes''.
    
    \item \textbf{Directional Ablation}: Given a true statement, we remove the component along $\mathbf{r}^{(l)}$ from the residual stream. If $\mathbf{r}^{(l)}$ encodes truth, the model's output should flip from ``Yes'' to ``No''.
\end{itemize}

\paragraph{Datasets.}
We use three domain-specific factual datasets, each consisting of a large number of true and false statements. The \textit{cities} dataset are from \citet{marks2023geometry} and \textit{element\_symb} and  \textit{animals\_class} datasets are subsets of datasets curated by \citet{azaria2023internal}. All statements are unambiguous and curated to evaluate world knowledge:

\subsection{Loss-Guided Concept Cone Discovery}

To discover a set of \textit{orthonormal basis vectors} that span a cone encoding the concept of truth, we optimize a composite loss that encourages each vector to:
\begin{enumerate}
    \item Induce truth behavior when added to false prompts.
    \item Inhibit truth behavior when ablated from true prompts.
    \item Preserve unrelated model behavior (i.e., maintain fidelity to non-targeted inputs).
\end{enumerate}

\paragraph{Objective.}
Following \citet{wollschlager2025geometry}, our optimisation target is a
three–term loss
\[
\mathcal{L}_{\text{total}}
  = \lambda_1\mathcal{L}_{\text{add}}
  + \lambda_2\mathcal{L}_{\text{ablate}}
  + \lambda_3\mathcal{L}_{\text{retain}},
\]
but with two implementation tweaks that adapt it to binary
truth–judgement:

\begin{enumerate}

\item \textbf{Binary generation.}  
      At generation time we zero out every logit except the two tokens
      \texttt{Yes} and \texttt{No} and sample \emph{one} token
      ($t{=}1$), which converts the addition/ablation terms into
      standard binary cross‐entropy losses.
\item \textbf{Wide-scope retention.}  
      To guard against collateral drift, \(\mathcal{L}_{\text{retain}}\)
      is measured on 30-token continuations of Alpaca instructions,
      providing a broad behavioural footprint.
\end{enumerate}

\medskip
\noindent
Formally the three components are:
\small{
\begin{definition}
\label{loss_add}
\[
    \mathcal{L}_{\text{add}} 
= -\frac{1}{|\mathcal{D}_{\text{false}}|} 
   \sum_{x\in\mathcal{D}_{\text{false}}} 
   \log \hat{y}_{\text{add}}(x + \mathbf{v})
   \tag{Add, target $y{=}1$} \\[3pt]
\]
\end{definition}
\begin{definition}
\[
    \mathcal{L}_{\text{ablate}} 
= -\frac{1}{|\mathcal{D}_{\text{true}}|} 
   \sum_{x\in\mathcal{D}_{\text{true}}} 
   \log \big[ 1 - \hat{y}_{\text{ablate}}(
   x - \mathbf{v} \mathbf{v}^\top x) \big]
   \tag{Ablate, target $y{=}0$} \\[3pt]
\]
\end{definition}

\begin{definition}
\[
\mathcal{L}_{\text{retain}} 
= \frac{1}{|\mathcal{D}_{\text{alpaca}}|} \sum_{x\in\mathcal{D}_{\text{alpaca}}}\! \!\!\!
D_{\mathrm{KL}} \big( 
   p_{0}(y_{1{:}30} \mid x)\big\|\
   p_{\mathbf{v}}(y_{1{:}30} \mid x) 
   \big)
   \tag{Retain, KL}
\]
\end{definition}
}

\normalsize
\noindent
Here \(\hat y_{\text{add}}\) and \(\hat y_{\text{ablate}}\) are the
post-softmax probabilities of outputting \texttt{Yes} after, respectively,
adding or ablating the truth vector \(\mathbf v\) at the chosen layer;
\(p_{0}\) and \(p_{\mathbf v}\) denote the unmodified and perturbed
30-token distributions for Alpaca prompts.  The scalars
\(\lambda_{1:3}\) balance steering power (\(\mathcal{L}_{\text{add}},
\mathcal{L}_{\text{ablate}}\)) against fidelity
(\(\mathcal{L}_{\text{retain}}\)).

\paragraph{Algorithm.}
\label{procedure}
We perform the following to generate a concept cone given the model, number of dimensions, layer and token position:
\begin{enumerate}
    \item Generate addition, ablation and retain targets for loss computation - 'Yes', 'No', and base model outputs on Alpaca training set respectively.
    \item Restrict model output logits to the tokens ``Yes'' and ``No'' to force binary classification .
    \item Generate a k-dimensional orthonormal basis for the cone using the concept cone method specified in \cite{wollschlager2025geometry} with our loss and model.
    \item Apply \textbf{activation addition} to false prompts and \textbf{directional ablation} to true prompts, targeting all the residual stream layers of the model
    \item Record the model’s response before and after intervention to determine if the answer flips to the desired label.
\end{enumerate}

\paragraph{Monte-Carlo sampling for Testing} 
After constructing the orthonormal basis vectors that define the truth cone, we aim to evaluate whether arbitrary directions within the cone also reliably mediate truthful behavior. Since any nonnegative linear combination of the basis vectors lies within the cone, we sample random directions by generating 64 sets of nonnegative coefficients $\{\lambda_1, \lambda_2, \dots, \lambda_k\}$ from a uniform distribution and normalizing the resulting vectors. Specifically, for each sample, we compute $\mathbf{v} = \sum_{i=1}^k \lambda_i \mathbf{b}_i$. We perform interventions on each of the sampled vectors to approximate the effectiveness full distribution of directions within the cone.

\begin{figure*}[t]
\centering
\includegraphics[width=\textwidth]{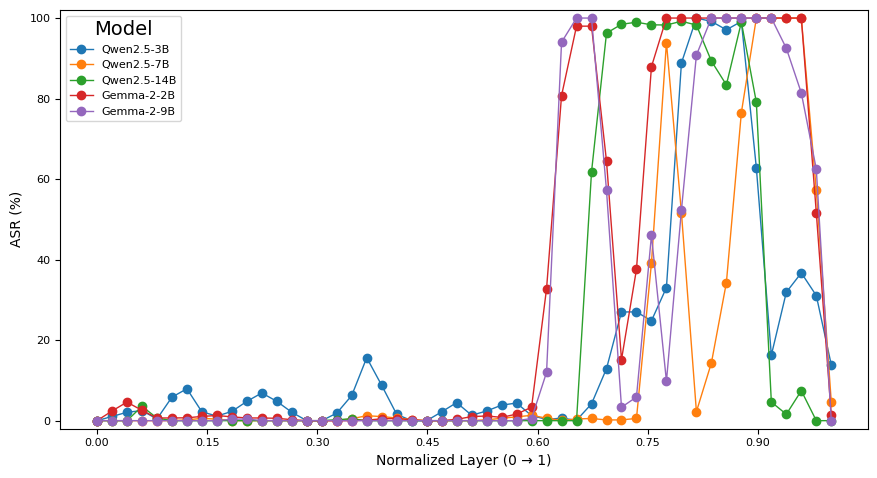}
\caption{The Attack Success Rate (ASR) of one dimensional cones across layers for Qwen and Gemma models. The layer numbers have been normalized across larger and smaller models. The effectiveness spikes rapidly in all models in the 0.60-0.75 range of normalized layer numbers.}
\label{fig:asr_layers}
\end{figure*}

\section{Experiments}

\subsection{Experiment 1: Localizing Truth Behavior in Layers and Token Positions}
\label{subsec:experiment1}

\paragraph{Goal.} We investigate which layers and token positions are most effective for capturing truth-related behavior. Since a linear direction is simply a one-dimensional concept cone, we first evaluate whether truth can be causally mediated at each layer using a single direction. If a model fails to encode truth behavior in a linear subspace at a given layer, it is unlikely that a higher-dimensional cone would succeed there either.

\paragraph{Procedure.} To do this, we train a one-dimensional cone (i.e., a linear direction) at each layer and across the last five token positions, and evaluate its \textit{Answer Switching Rate} (ASR). Specifically, we measure the success of activation-based interventions across multiple datasets and model families by computing the ASR -- the proportion of inputs which affect model outputs after an intervention.

Formally, we define the \textbf{Answer Switching Rate (ASR)} as:

\textbf{Definition of ASR}
\[
\text{ASR} =
\frac{
    \begin{array}{c}
        \# \text{ of prompts whose output} \\
        \text{becomes untruthful after ablation}
    \end{array}
}{
    \begin{array}{cc}
         \text{baseline }\# \text{ of prompts that the model} \\
        \text{ answers truthfully}
    \end{array}
}
\]

In practice, the baseline is almost always the same as the total number of prompts as the models nearly always achieve full accuracy when answering our simple propositions.

\paragraph{Results.} Across both model families and sizes, we find that truth-related directions reliably emerge in the middle to later layers (specifically, between 60–75 percent of the normalized layer depth). As shown in Figure~\ref{fig:asr_layers}, ASR increases sharply in this range before decreasing sharply again in the very last layers. Additionally, we find that the final token position consistently yields the strongest interventions, consistent with prior work showing that high-level decisions often accumulate at the end of the sequence \cite{arditi2024refusal, burger2024truth}.

Based on these findings, we restrict our concept cone search to this high-performing region of the network. This choice is motivated both by empirical signal strength and by computational efficiency.

\begin{figure*}[t]
\centering
\includegraphics[width=\textwidth]{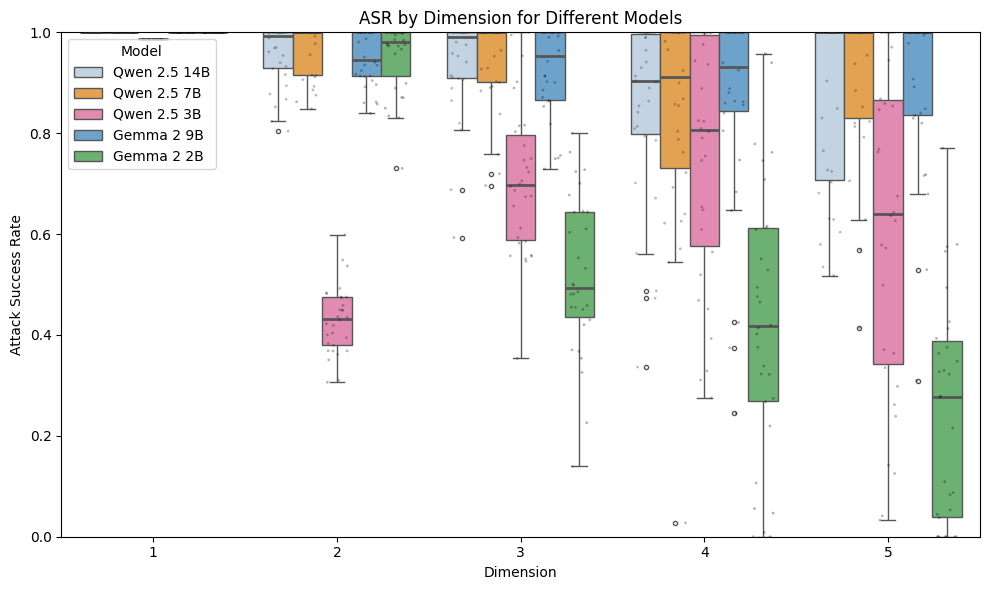}
\caption{The Answer Switching Rate (ASR) of cones from dimensions 1 to 5 across Qwen2.5 and Gemma2 models with boxplots showing the Monte Carlo sampling.}
\label{fig:asr_dims}
\end{figure*}

\subsection{Experiment 2: Truthfulness Steering Across Models and Dimensions with Cones}
\label{subsec:experiment2}

\paragraph{Goal.}
The aim of this experiment is to assess the effect of increasing dimensionality on the ability causally mediate truthful behavior. While previous results show that a single direction can causally influence truthfulness, we seek to determine how many additional, orthogonal directions can also support this behavior before unrelated features begin to dilute the effect. We do this across multiple models from the Qwen-2.5 and Gemma-2 families, spanning a range of parameter sizes.

\paragraph{Procedure.}
For each model, we construct concept cones with dimensionalities ranging from 1 to 5. Each cone is generated using the optimization procedure described in Section~\ref{Sec: Methodology}, which ensures the basis vectors satisfy both causal and retention constraints. To evaluate generalization across the cone space, we perform Monte Carlo sampling within each cone by drawing random nonnegative combinations of the basis vectors. We then measure the effectiveness of each sampled direction using ASR defined previously.

\paragraph{Results}

Table~\ref{tab:asr-results} presents the Answer Switching Rate (ASR) across five language models (Qwen2.5-3B, Qwen2.5-7B, Qwen2.5-14B, Gemma-2-2B, and Gemma-2-9B) as a function of the dimensionality of the concept cone used for intervention. Each ASR value reflects the average success rate across Monte Carlo samples drawn from the cone of that dimension.

\begin{table}[h]
\centering
\small{\small{
\caption{Answer Switching Rate after intervention across models and cone dimensions.}
\label{tab:asr-results}
\begin{tabular}{lccccc}
\toprule
\textbf{Model}      & \textbf{1}(DIM) & \textbf{2}    & \textbf{3}    & \textbf{4}    & \textbf{5}    \\
\midrule
Qwen 14B & 100   & 100  & 98.6  & 91.2  & 100   \\
Qwen 7B  & 100   & 100   & 100   & 100   & 100   \\
Qwen 3B  & 98.6  & 45.1  & 67.2  & 78.9  & 65.3  \\
Gemma 9B   & 100   & 100   & 100   & 98.6  & 97.3  \\
Gemma 2B   & 100   & 100   & 53.7  & 43.1  & 27.1  \\
\bottomrule
\end{tabular}
}}
\end{table}

\paragraph{Interpretation.}
The results in Table~\ref{tab:asr-results} suggest that increasing the dimensionality of the concept cone generally improves the model’s ability to internalize and respond to truth-aligned interventions. Larger models, such as Qwen-2.5-7B and Gemma-2-9B, maintain high ASR even as dimensionality increases, meaning that higher dimension cones exist within their activation space. This is consistent with findings from \citet{wollschlager2025geometry} in the domain of refusal behavior.

However, the trend is not monotonic: beyond a certain point, ASR begins to decline, indicating that additional directions may start to capture unrelated features and dilute the effectiveness of the intervention. This effect is especially evident in smaller models, where cone dimensions above 2 or 3 yield diminishing or negative returns. Nonetheless, the models still show multiple dimensions that independently support truth-aligned behavior. For example, both Qwen-7B and Gemma-9B maintain near-100\% ASR across all tested dimensions, showing that there is at least a 5-dimensional cone that causally mediates truth.

\subsection{Experiment 3: Retention of General Capabilities via KL Divergence}
\label{subsec:experiment2}

\paragraph{Goal.}
While the purpose of truth-direction interventions is to modify the model’s factual response behavior, we must ensure they do not interfere with unrelated capabilities. This experiment evaluates how much the intervention alters model output on a general instruction-following benchmark, using KL divergence as a metric of deviation. This operationalizes the $\mathcal{L}_{\text{retain}}$ loss term defined in Section \ref{Sec: Methodology}.

\paragraph{Dataset.}
We use the \textit{ALPACA} \cite{taori2023alpaca} dataset, a popular instruction-following benchmark designed to elicit helpful, safe, and general-purpose completions.

\paragraph{Procedure.}
We randomly select 200 prompts that are unlikely to invoke factual disputes (e.g., summarization, rewriting, math, or basic instructions).

For each cone that we generate, we compare the original model’s outputs to those produced after applying directional ablation using the discovered truth directions. The intervention is applied globally (all tokens, all layers) as described in Equation~\ref{eq:projection}. As a threshold, we don't consider cones with basis vectors with a KL Divergence above 0.1, following the precedent of papers like \cite{arditi2024refusal}.

\paragraph{Results.}
We report the mean KL divergence across 200 Alpaca prompts in Table~\ref{tab:kl-alpaca}. We find that the truth-direction ablation leads to only minimal divergence from the original output distribution, suggesting that the intervention does not significantly affect unrelated capabilities. 

\begin{table}[h]
\centering
\caption{Mean KL divergence on Alpaca prompts (lower is better).}
\label{tab:kl-alpaca}
\begin{tabular}{lc}
\toprule
\textbf{Model} & \textbf{Mean KL Divergence} \\
\midrule
Qwen2.5-14B & 0.038 \\
Gemma-2-2B   & 0.045 \\
Qwen2-7B   & 0.026 \\
Gemma-2-9B   & 0.031 \\
\bottomrule
\end{tabular}
\end{table}

\paragraph{Interpretation.}
All models show low average KL divergence, especially the larger variants. This suggests that the discovered truth directions are highly specific and do not interfere with general instruction-following behavior. The effectiveness of $\mathcal{L}_{\text{retain}}$ as a regularization objective is empirically supported by this result.

\subsection{Experiment 4: DIM vs.\ Cone Alignment}
\label{subsec:exp3}
We measure how closely the classic DIM truth
vector aligns with the orthonormal directions discovered by our
concept cone.  Cosine similarity is reported in
Table~\ref{tab:cos-sim}; values near 1 indicate strong overlap.

\begingroup
  \setlength{\tabcolsep}{3pt} 
  \renewcommand{\arraystretch}{0.8} 
  \setlength{\abovecaptionskip}{3pt}
  \setlength{\belowcaptionskip}{3pt}

  \begin{table}[h]
    \centering\scriptsize
    \caption{Cosine similarities between the DIM direction and cone basis vectors in Gemma-2-9B, transposed for dimensions 2–5.}
    \label{tab:cos-sim}
    \begin{tabular}{@{}lcccc@{}}
      \hline
       & \textbf{Dim 2} & \textbf{Dim 3} & \textbf{Dim 4} & \textbf{Dim 5} \\
      \hline
      $v_{1}$ & $1.23\times10^{-1}$   & $1.45\times10^{-1}$   & $2.00\times10^{-1}$   & $2.26\times10^{-1}$    \\
      $v_{2}$ & $-3.72\times10^{-9}$  & $1.74\times10^{-9}$   & $3.03\times10^{-9}$    & $-6.98\times10^{-10}$  \\
      $v_{3}$ & —                     & $1.16\times10^{-9}$   & $-2.33\times10^{-9}$   & $-4.19\times10^{-9}$   \\
      $v_{4}$ & —                     & —                     & $2.33\times10^{-10}$    & $8.38\times10^{-9}$    \\
      $v_{5}$ & —                     & —                     & —                      & $3.03\times10^{-9}$    \\
      \hline
    \end{tabular}
  \end{table}
\endgroup

\paragraph{Results.}
Only the first cone axis has any alignment with DIM, confirming that DIM captures just one facet of the multi-dimensional truth subspace; the remaining axes encode additional, orthogonal structure.

\section{Discussion}
Our findings reveal that while a single direction derived from the DIM method already captures a strong causal representation of truth in LLMs, it does not fully exhaust the structure underlying truth-related behavior. Through our concept cone approach, we identified additional orthogonal directions with low cosine similarity to the DIM vector that also reliably steer model outputs on propositional truth tasks. This suggests that truthful behavior may not be confined to a single axis—multiple directions and can be independently influenced. These directions likely correspond to distinct or semantically adjacent components of factual reasoning, such as modality, certainty, or domain-specific features.

The success of both DIM and cone-based interventions suggests that truth may be linearly separable in the model’s representation space. While the directions that independently modulate truthful behavior can imply that this structure may be richer than a single linear axis, it does not necessarily prove that the underlying representation of truth is nonlinear. It does, however, open up important questions in the context of model deception, robustness, and interpretability. If multiple, semantically adjacent directions can influence truthfulness, models may be more vulnerable to manipulations that subtly shift their outputs without obvious signs of tampering within the first truth direction. Understanding the geometry of these truth-related subspaces is essential for building models that are not only aligned, but resilient to adversarial or unintended shifts in behavior.

\section{Conclusion and Future Directions}
In this work, we showed that multi-dimensional concept cones can reliably steer the behavior of LLMs on true/false propositions across multiple architectures and sizes, while minimally impacting unrelated behaviors. Our results reveal that, beyond a single “truth direction,” there exists a robust subspace of activation vectors whose positive combinations consistently modulate factuality. These findings show increasing promise for concept cones as an interpretability toolkit and underscore new avenues and risks for alignment, calibration, and adversarial manipulation of model truthfulness.

Several promising avenues remain. Concept cone search reliably uncovers a subspace, but we are yet to find semantically meaningful labels for the basis vectors. Future work could pair cones with automated clustering or sparse autoencoding so that each basis vector corresponds to an interpretable facet of truth (e.g. temporal facts, geographic facts, or commonsense). Extending the method to larger, instruction-tuned models and to multimodal settings would also test its robustness and reveal whether these semantic dimensions persist across scale and modality.

\section{Limitations}
\subsection{Model Scale}
All experiments were conducted on relatively small open-source models (1.5B–7B parameters). While we observe clear directional structure in the residual stream of mid-sized models, these findings may not generalize to larger frontier models or architectures with substantially different alignment protocols. Notably, smaller models exhibit lower ASR, with PCA visualizations revealing weaker separation of truth-related directions, especially in later layers. This suggests that representational abstraction of truth may emerge more clearly with scale.

\subsection{Scope}
Our operationalization of truth/factfulness is deliberately narrow, limited to simple unambiguous propositional facts that have a clear true or false answer. While this allowed for clean experimental design, it does not capture more complex notions of truth that may be more widely applicable, such as context-dependent claims or statements that involve some kind of subjectivity. As a result, the discovered directions or cones may not generalize to broader or more nuanced conceptions of truth. 

Our experiments span only two model families (Gemma and Qwen). It remains an open question whether the discovered directions are robust to architectural variation or fine-tuning differences. Evaluating cross-family generalization, especially to models trained with stronger alignment (e.g., RLHF or human preference tuning), is an important direction for future work.

\subsection{Subspace Understanding}
Although we demonstrate that a low-dimensional subspace (or “cone”) can causally mediate truth behavior, our method does not guarantee the discovery of a maximally informative or interpretable subspace. We leave for future work the development of principled methods – for example using sparsity constraints, disentanglement metrics, or unsupervised clustering – to assign semantic meaning to individual cone axes.

\bibliographystyle{plain}
\bibliography{references}

\newpage
\appendix
\label{sec:appendix}
\onecolumn

\section{Setup Details}
\label{setup-details}

\subsection{Implementation Details}
\label{sec:impl}

Experiments run on NVIDIA H100-80GB GPUs using \texttt{PyTorch} 2.20 and \texttt{HF Transformers} 4.41. Core settings are in Table \ref{tab:impl}.

\begin{table}[h]
\centering
\caption{Hardware and hyper-parameters.}
\label{tab:impl}
\begin{tabular}{@{}ll@{}}
\hline
GPUs           & 1×H100 (probing)                \\
Batch size     & 4                                  \\
Number of Samples (during training) & 16            \\
Precision      & bfloat16                           \\
Optimizer      & AdamW \\
Code base      & \texttt{transformer-lens 0.9.1} \\
 & \texttt{nnsight 0.3.7} \\
\hline
\end{tabular}
\end{table}

\subsection{Datasets}

\begin{table}[h]
    \centering
    \caption{Full datasets used in experiments}
    \label{tab:datasets}
    \begin{tabular}{@{}ll@{}}
    \hline
    Cities &  \href{https://github.com/sciai-lab/Truth_is_Universal/blob/main/datasets/cities.csv}{Link}\\
    Animals &  \href{https://github.com/sciai-lab/Truth_is_Universal/blob/main/datasets/animal_class.csv}{Link}\\
    Elements & \href{https://github.com/sciai-lab/Truth_is_Universal/blob/main/datasets/element_symb.csv}{Link}\\
    \hline
    \end{tabular}
\end{table}

\subsection{Models}

\begin{table}[h]
    \centering
    \caption{All Models used in Experiments}
    \label{tab:models}
    \begin{tabular}{@{}ll@{}}
    \hline
    Qwen2.5-3B-Instruct &  \href{https://huggingface.co/Qwen/Qwen2.5-3B-Instruct}{Link}\\
    Qwen2.5-7B-Instruct & \href{https://huggingface.co/Qwen/Qwen2.5-7B-Instruct}{Link}\\
    Qwen2.5-14B-Instruct & \href{https://huggingface.co/Qwen/Qwen2.5-14B-Instruct}{Link}\\\
    Gemma-2-2B-IT & \href{https://huggingface.co/google/gemma-2-2b-it}{Link}\\
    Gemma-2-9B-IT & \href{https://huggingface.co/google/gemma-2-9b-it}{Link}\\
    \hline
    \end{tabular}
\end{table}

\section{Additional Experiments}
\label{additional-experiments}
\subsection{Sentiment}

Previous literature \cite{tigges2023linear} suggests that sentiment has a linear representation, similar to other concepts such as refusal \cite{arditi2024refusal}. We tried to extend our methodology to sentiment to determine whether it has a concept cone representation. In particular, we trained a concept cone on the Stanford Sentiment Treebank \cite{socher-etal-2013-recursive} which consists of 10,662 one-sentence movie reviews with fully labeled parse trees. We failed to find a meaningful concept cone for sentiment. Further work could explore alternative techniques for finding a higher-dimensional representation for sentiment.

\subsection{Toxicity}

We explored the existence of higher-dimensional representations for toxicity. We trained a concept cone on the ToxiGen benchmark \cite{hartvigsen2022toxigen} which consists of 274,000 toxic and benign phrases about minority groups such as such as Black, Asian, Latino, Jewish, Muslim, LGBTQ+, women, and individuals with mental or physical disabilities. We first computed a difference-in-means (DIM) direction. We also computed a direction by taking the difference between the activations of a student model that generates toxic output (typically an instruction-tuned model) and a teacher model that doesn't generate toxic output (typically an instruction-tuned model) on the dataset, a method described in \citet{wang2023trojan}. Testing this direction using ablation failed, as the resulting output was unintelligible. Testing the DIM direction using ablation did disable strong refusal (the model did not say "I cannot respond" outright) but would still refuse to generate toxic output. Since we were unable to obtain a valid linear direction for toxicity, we were unable to generate high-quality targets and as a result, were unable to train a valid cone. Further work could explore better methods for obtaining a direction for toxicity.

\section{Full List of System Prompts}
\label{system-prompts}
System-level instructions exert a strong influence over an LLM’s eventual answer:  
recent work shows that even minor wording differences can sway a model toward a
different multiple-choice option, particularly when those differences encode
authority or confidence \cite{anagnostidis2024susceptible}; that prompts which
explicitly acknowledge uncertainty often elicit more calibrated responses
\cite{zhou2025uncertainty}; and that a model’s overall “steerability” rises or
falls with the directness of the request phrasing \cite{miehling2025steerability}.

\vspace{0.6em}
\begin{table}[h]
\centering
\caption{Example system prompts that force a binary \texttt{Yes}/\texttt{No} answer.}
\label{tab:yn-prompts}
\begin{tabular}{cl}
\hline
\# & \textbf{Prompt wording} \\
\hline
1 & Respond to the following statement with a \textbf{“Yes”} or a \textbf{“No.”} \\[2pt]
2 & Respond to the following statements with either \textbf{“Yes”} or \textbf{“No”} based on their factual accuracy. \\[2pt]
3 & Answer with \textbf{exactly one word}—\textbf{"Yes"} if the statement is true, otherwise \textbf{"No"}. \\[2pt]
4 & Based on your best estimate of the following statement, respond \textbf{“Yes”} or \textbf{“No.”}\\[2pt]
\hline
\end{tabular}
\end{table}

\section{Token choice to represent True/False}
\label{token-choice}
In our experiments, we frame factual statements as binary questions and constrain the model’s output to the tokens \texttt{Yes} and \texttt{No}. We also tested alternative tokenizations such as \texttt{yes}/\texttt{no} and \texttt{true}/\texttt{false}, but found that these variations had no significant effect on steering outcomes or ASR. Interestingly, when the output vocabulary is left unrestricted, models occasionally respond in non-English equivalents of “Yes” and “No” (e.g., \texttt{Sí}, \texttt{Nein}) following truth-direction interventions. This suggests that the underlying truth representation may generalize across lexical choices, although further investigation is needed to confirm cross-lingual consistency.

\section{Cosine Similarities}
\label{cos-similarities}

We see the same trend for cosine similarity across models, where other than the first dimension, all increasing dimensions have extremely low cosine similarity to the DIM direction.

\begin{table}[h]
  \centering
  \caption{Cosine similarities between the DIM direction and cone basis vectors in Qwen-2.5-9B, transposed for dimensions 2–5.}
  \label{tab:cos-sim-appendix}
  \begin{tabular}{@{}lcccc@{}}
    \hline
     & \textbf{Dim 2} & \textbf{Dim 3} & \textbf{Dim 4} & \textbf{Dim 5} \\
    \hline
    $v_{1}$ & $-1.57\times10^{-1}$   & $1.82\times10^{-1}$   & $1.34\times10^{-1}$   & $1.67\times10^{-1}$    \\
    $v_{2}$ & $-4.23\times10^{-9}$  & $2.91\times10^{-9}$   & $-1.08\times10^{-9}$   & $-5.74\times10^{-10}$  \\
    $v_{3}$ & —                     & $3.56\times10^{-9}$   & $-7.42\times10^{-9}$   & $6.13\times10^{-9}$    \\
    $v_{4}$ & —                     & —                     & $2.87\times10^{-9}$   & $-2.45\times10^{-10}$   \\
    $v_{5}$ & —                     & —                     & —                      & $4.39\times10^{-9}$    \\
    \hline
  \end{tabular}
\end{table}

\section{Code}
All code will be open-sourced on Github.

\section{PCA Visualizations}
As a cursory introduction into understanding literature of linear representations of truth, we recreated Principal Component Analysis visualizations of all models used in the experiments on the datasets onto their top two principal components. All components are listed below.

\begin{figure*}[h]
\centering
\includegraphics[width=\textwidth]{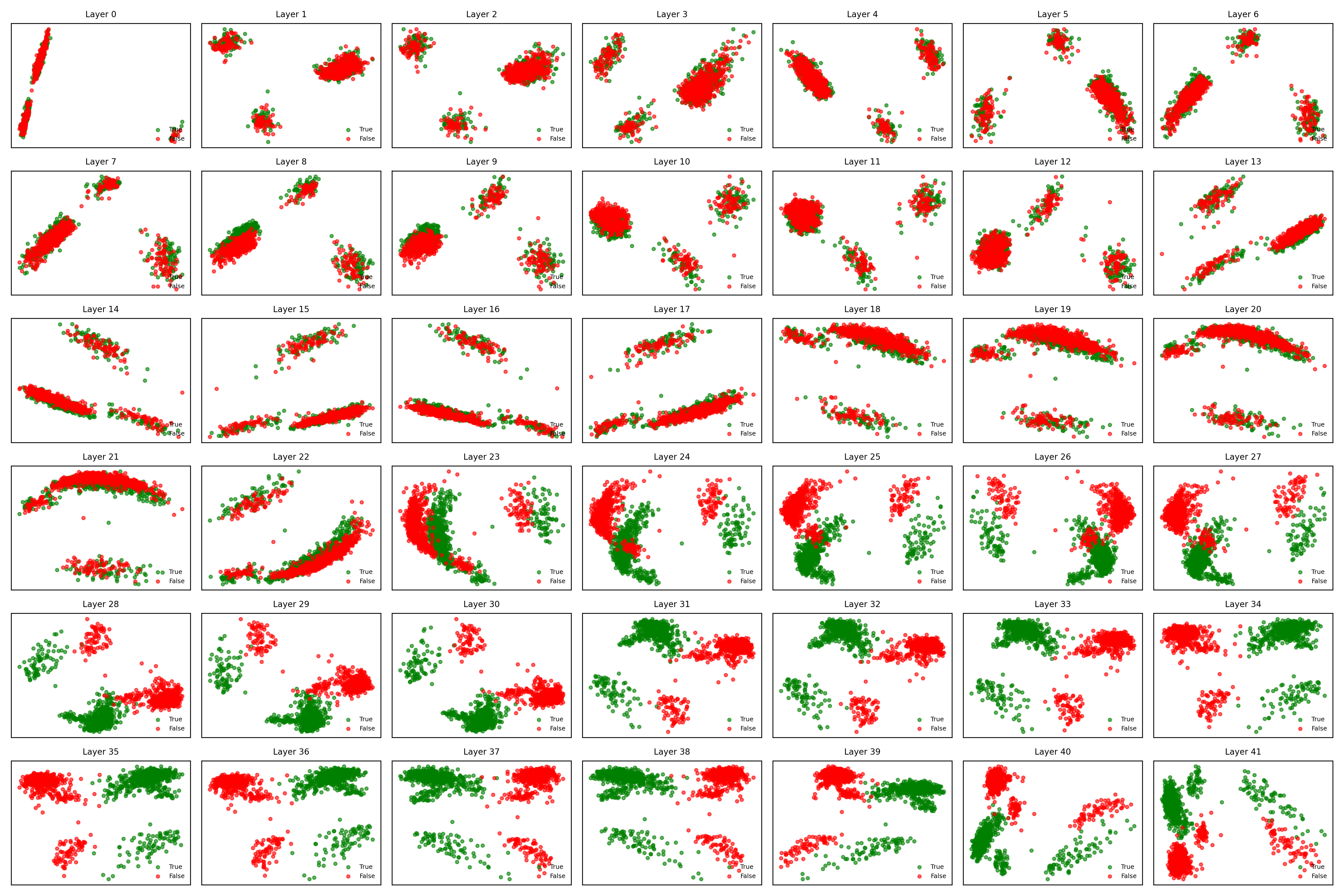}
\caption{Projections of Gemma-2-9B, representations of datasets onto their top two PCs, across all layers.}
\label{fig:pca}
\end{figure*}

\begin{figure*}[h]
\centering
\includegraphics[width=\textwidth]{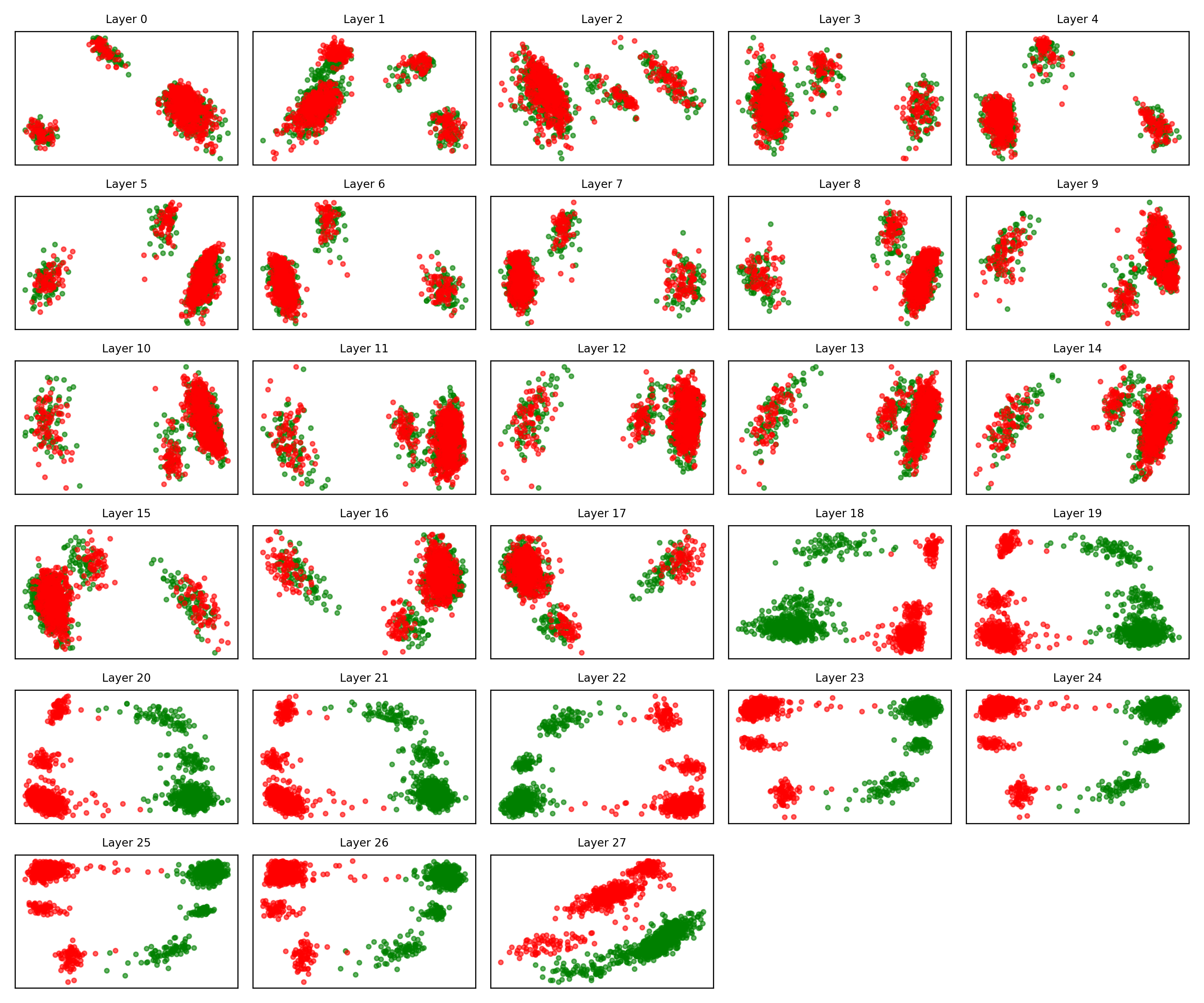}
\caption{Projections of Qwen2.5-7B representations of datasets onto their top two PCs, across all layers.}
\label{fig:pca}
\end{figure*}

\end{document}